		\documentclass[final,1p,times]{elsarticle}
		
		\usepackage{amsmath, amsfonts, amssymb}
		\usepackage{geometry}
		\usepackage{booktabs}
		\usepackage{float}
		
		\journal{review}
		
\begin{document}
		
		\begin{frontmatter}
		
		
		
		\title{Growing Cosine Unit: A Novel Oscillatory Activation Function That Can Speedup Training and Reduce
Parameters in Convolutional Neural Networks}
		
		
		\author [label1]{Mathew Mithra Noel}
		\author [label2]{Arunkumar L}
		\author [label3]{Advait Trivedi} 
		\author [label4]{Praneet Dutta }

		\affiliation[label1]{organization={Vellore Institute of Technology}, addressline={mathew.mithra@gmail.com}, country={India}}
		
		\affiliation[label2]{organization={Vellore Institute of Technology}, addressline={arunk609@gmail.com}, country={India}}
		
		\affiliation[label3]{organization={Red Hat Inc.}, addressline={astrived@ncsu.edu}, country={USA}}
		
		\affiliation[label4]{organization={Independent researcher}, addressline={praneetd@alumni.cmu.edu}, country={USA}}

		\begin{abstract}

		Convolutional neural networks have been  successful in solving many socially important and economically significant problems. This ability to learn complex high-dimensional  functions hierarchically can be attributed to the use of nonlinear activation functions. A key discovery that made training deep networks feasible was the adoption of the Rectified Linear Unit (ReLU) activation function to alleviate the vanishing gradient problem caused by using saturating activation functions. Since then, many improved variants of the ReLU activation have been proposed. However, a majority of activation functions used today are non-oscillatory and monotonically increasing due to their biological plausibility. This paper demonstrates that oscillatory activation functions can improve gradient flow and reduce network size. Two theorems on limits of non-oscillatory activation functions are presented.  A new oscillatory activation function called Growing Cosine Unit(GCU) defined as $ C(z) = z \cdot \cos z$ that outperforms Sigmoids, Swish, Mish and ReLU on a variety of architectures and benchmarks is presented. The GCU activation has multiple zeros enabling single GCU neurons to have multiple hyperplanes in the decision boundary. This allows single GCU neurons to learn the XOR function without feature engineering.  Experimental results indicate that replacing the activation function in the convolution layers with the GCU activation function significantly improves performance on CIFAR-10, CIFAR-100 and Imagenette.
		
		\end{abstract}

%		\begin{graphicalabstract}
%		
%		\includegraphics [width = 5 cm]{relu_layer10}
%		\includegraphics [width = 5 cm]{xcosx_layer10} \\
%		\includegraphics [width = 5 cm]{activation_functions}
%		\includegraphics [width = 6 cm]{XOR_GATE}
%		\end{graphicalabstract}
%		
%		%%Research highlights
%		\begin{highlights}
%		\item Proposed activation function outperforms all popular activation functions
%		\item Two theorems on limits of non-oscillatory activation functions are presented
%		\item A single neuron solution to the famous XOR problem is presented for the first time
%		\item Oscillatory activation functions are shown to improve gradient flow and accuracy
%		\item Oscillatory activation functions are shown to result in smaller networks
%		\end{highlights}
%		

		
		\end{frontmatter}
		
		
		\section{Introduction}

		The quintessential feature of deep Convolutional Neural Networks (CNNs) is their ability to learn arbitrarily complex nonlinear mappings between high-dimensional Euclidean spaces \cite{hornik1989multilayer} \cite{leshno1993multilayer}.  This  universal approximation feature (\cite{scarselli1998universal}, \cite{392253}, \cite{cybenko1989approximation}, \cite{funahashi1989approximate}) is  critically  dependent  on  the  nature  of  the  activation  function non-linearity used in each layer of the neural network. Training a neural network might be viewed as adjusting a set of parameters to scale, compress, dilate, combine and compose simple nonlinear activation functions to approximate the complex nonlinear target function. Thus, the use of more complex activation functions might allow the target function nonlinearity to be approximated using fewer neurons. This paper proposes an oscillating activation function called Growing Cosine Unit (GCU) that outperforms all known activation functions on benchmark datasets and allows function approximation tasks to be performed with fewer neurons. For example, the famous XOR problem can be solved with a single GCU neuron instead of 3 sigmoidal neurons \cite{sprinkhuizen1996error}. The GCU activation function has zeros only at isolated points and hence overcomes the \textquotedblleft neuron death problem\textquotedblright where the output of ReLU neurons get trapped at zero \cite{douglas2018relu}.
		
		The nonlinearity of the activation function  is essential, since the composition of any finite number of linear functions is equivalent to a single linear function. Hence any network however large but composed of purely linear neurons is equivalent to a single layer of linear neurons. Further networks of neurons with the linear activation function ($ g(z) = z $) are limited to solving linearly separable classification problems. Despite the critical importance of the nature of the activation function in determining the performance of  neural  networks, simple  monotonic non decreasing nonlinear  activation  functions  are universally  used. In  this  paper, we  explore  the  potential benefits  of  using oscillatory   nonlinear   activation   functions   in   deep   neural   networks.

		In the past, sigmoidal saturating activation functions were widely used because these functions approximate the step or signum functions (used in Rosenblatt's  perceptron \cite{4066017}) while  still being differentiable \cite{DBLP:journals/corr/abs-1811-03378}. The outputs of s-shaped saturating activations have the important property of being interpretable as a binary yes/no decision and hence are useful. However, deep neural networks composed of purely sigmoidal activation functions are hard to train, due to the vanishing gradient phenomenon which arises when saturating activation functions are used.  The adoption of the non-sigmoidal Recti-Linear Unit (ReLU) \cite{DBLP:journals/corr/abs-1803-08375} activation function to alleviate the vanishing gradient problem is considered a milestone in the evolution of deep neural networks \cite{poirazi2003pyramidal} \cite{agarap2018deep}.

		During training with the Backpropagation algorithm, the  parameters of a network are continually updated in the direction of the negative gradient \cite{rumelhart1986learning}. Hence small gradients lead to stagnation in learning and slow parameter updates. The derivative of sigmoidal activation functions is small outside a small closed interval around zero (usually [-5 , 5]). In particular $ \exp (-5) < 0.01 $ and hence activation functions composed purely of exponentials, such as logistic-sigmoid and tan-sigmoid will saturate outside this narrow range.
		
		Furthermore, with uni-polar activation functions (functions that take purely non-negative values like logistic-sigmoid), the outputs of a layer can get combined to form large positive values leading to the saturation of neurons in the next layer. Thus, activation functions that do not shift the mean of the input towards positive or negative values (such as tanh(z)) reduce saturation of succeeding layers and hence perform better.

		In the past a wide variety of activation functions have been explored \cite{DBLP:journals/corr/abs-1811-03980}, \cite{DBLP:journals/corr/abs-1710-05941}. Past research indicates that activation functions that have larger derivative values for a wider set of input values perform better  \cite{pmlr-v9-glorot10a}. In particular the use of the ReLU like activation functions result in faster training compared to saturating sigmoidal type activation functions because these activation functions do not saturate for a wider range of inputs. 
		
		Some drawbacks of ReLU like activation functions:
		\begin{itemize}
			\item The derivative of the loss function $J$ with respect to the weight matrix $W^k$ of layer $k$ is $\frac{\partial J}{\partial W^k} = \delta^k(a^{k-1})^T$, where $a^{k-1}$ is the vector of activations of layer $k-1$ and $\delta^k$ is the vector of derivatives of the loss function with respect to the net weighted inputs. Thus if $a_i^{k-1}$ or $\delta_j^k$ is small then the derivative of the loss with respect to the weight $w_{ij}$ is also small and the weight $w_{ij}$ is not updated and learning stagnates.
			\item Bias Shift: There is a positive bias in the network for subsequent layers, as the mean activation is always greater than zero. Since the outputs of all ReLU units are non-negative the outputs can combine to produce very large positive inputs to subsequent layers farther away from the input leading to possible saturation and numerical accuracy issues.
			\item The delta for a particular layer is $\delta ^k = \frac{\partial J}{\partial{ z^k}} = ( (W^{k+1})^T \delta^{k+1} ) \bigodot g'(z^k)$, where $g'(z^k)$ is the derivative of the activation function. So ReLU like activation functions that have zero or small derivative for negative values result is small $\delta ^k$ values leading to stagnation in learning.
		\end{itemize}
		
		In recent years variants of ReLU like SELU \cite{DBLP:journals/corr/KlambauerUMH17} and ELU \cite{Clevert2016FastAA} have been successful to an extent in mitigating the above shortcomings. Swish \cite{Ramachandran2017SwishAS} and Mish \cite{DBLP:journals/corr/abs-1908-08681} represent a new class of non-monotonic functions that offer promising results across different benchmarks.

		Despite the popularity of a wide variety of activation functions and neural network architectures, all  networks suffer from a fundamental limitation in that individual neurons can exhibit only linear decision boundaries. Multilayer neural networks with nonlinear activations are needed to achieve nonlinear decision boundaries. This paper explores proposes a new oscillatory activation function that allows individual neurons to exhibit nonlinear decision boundaries thus removing a fundamental limitation of the classical  neuron model. 
		\\

		The main contributions of this work are:
		\begin{itemize}
		    \item A new activation function, Growing Cosine Unit (GCU) defined by $C(z)= z \cdot \cos (z)$  has been proposed. The advantages of using oscillatory activation functions to improve gradient flow and alleviate the vanish gradient problem has been demonstrated.

		     \item A  solution to the classic and long-standing XOR problem has been presented by successfully training a single GCU neuron to learn the XOR function without feature-engineering.
		    
		    \item Two theorems that characterize the limitation of certain class of activation functions are presented. 
		    
		     \item A comparison of the proposed GCU activation with popular activation functions on a variety of benchmark datasets is presented in Section \ref{section:results}. These experimental results clearly indicate that the GCU activation is computationally cheaper than the state-of-art Swish and Mish activation functions. The GCU activation also reduces training time and allows classification problems to be solved with smaller networks. 
		    
		 \end{itemize}
		
		\section{Oscillatory Activation functions}
		
		This paper explores the potential performance benefits and effects of  using  oscillatory activation functions in neural networks. In the past oscillatory and non-monotonic activation functions have been largely ignored. In the following, the famous XOR problem that requires a 2-layer network with 3 neurons (2 hidden and one output) is solved with a single GCU neuron. This example demonstrates the more powerful function approximation ability of the GCU neuron.

		\subsection{Learning the XOR function using a single neuron}
		The famous XOR problem is task of training a neural network to learn the XOR gate function. It was first pointed out by Papert and Minsky \cite{minsky69perceptrons}, \cite{brutzkus2019larger} that a single neuron cannot learn the XOR function since a single hyperplane (line in this case) cannot separate the two classes in the XOR dataset. This fundamental limitation of single neurons (or single layer networks) lead to pessimistic predictions for the future of neural network research and was responsible for a brief hiatus in the history of AI. This paper shows that this limitation in learning the XOR function does not apply to neurons with oscillating activation function having multiple zeros.

		In the past many attempts have been made to solve the XOR problem with less than 3 neurons. in \cite{nitta2003solving}, the XOR problem is solved with a single complex-valued neuron. In cite{gomez2006polynomial}, polynomial discrete time cellular neural networks to solve the XOR problem. In \cite{reljan2017solving}, the XOR problem is solved using spiking neural networks. Despite these attempts to solve the XOR problem with more complex neuronal models a simple single neuron solution using the classical neuron model has not been presented. 
		
		The XOR problem \cite{sprinkhuizen1996error} is the task of learning the following dataset:
		\begin{equation}\label{Eq: XOR dataset}
		    D = \left\{ 
		    (\begin{bmatrix} -1\\-1 \end{bmatrix}, -1), (\begin{bmatrix} 1\\-1 \end{bmatrix}, 1),
		    (\begin{bmatrix} -1\\1 \end{bmatrix}, 1),
		    (\begin{bmatrix}  1\\1 \end{bmatrix}, -1)
		    \right\}
		\end{equation}
		
		\begin{figure}[H]
		\center
		    \includegraphics[width=0.5\linewidth]{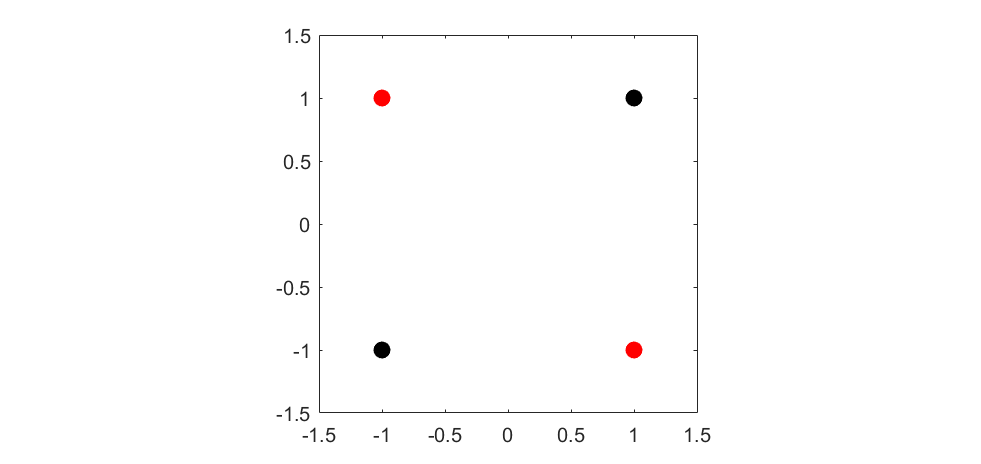}
		    \caption{The XOR problem is the task of learning the XOR function with the smallest network. The red dots must be classified as positive (represented by +1) and the black dots must be classified as negative (represented by -1). A single line (hyperplane) cannot separate the two classes.}
		     
		\end{figure}
		
		\begin{figure}[H]
		    \centering
		    \includegraphics[width=0.4\linewidth]{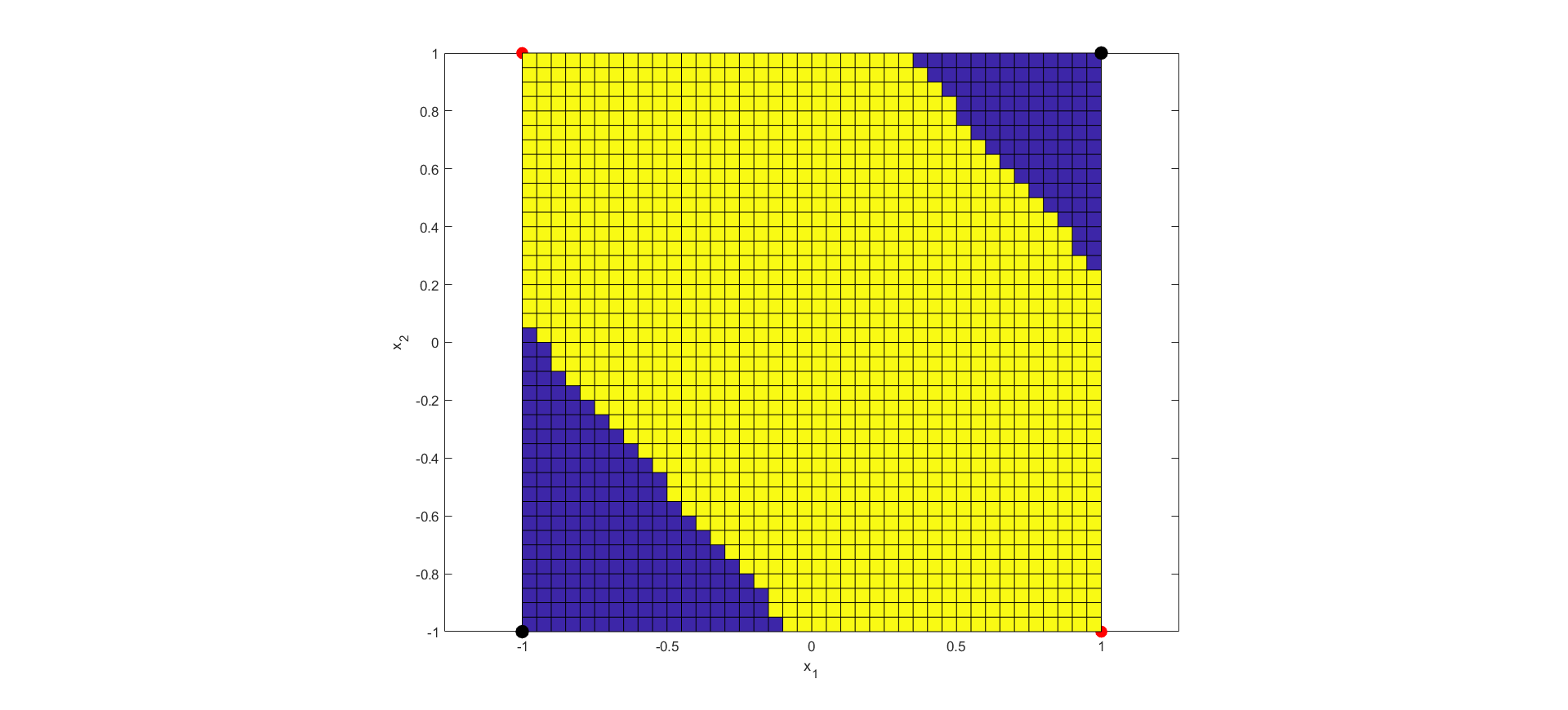}
		    \includegraphics[width=0.55\linewidth]{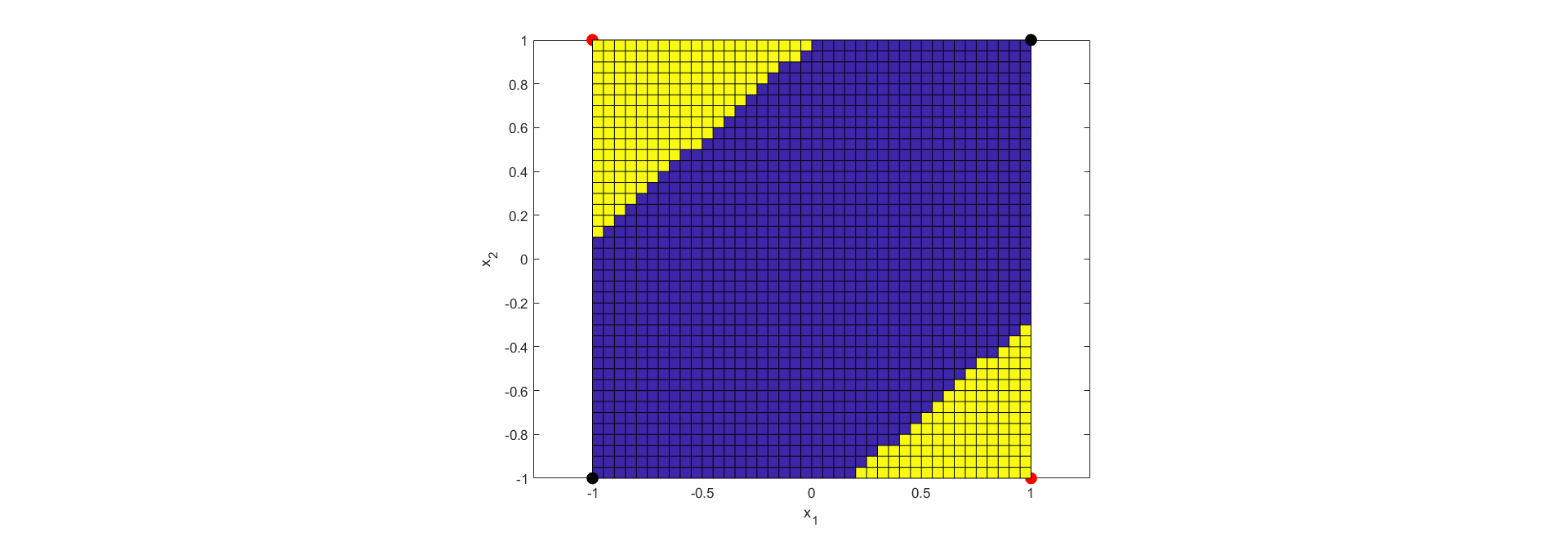}
		    \caption{Two solutions to the XOR problem learnt by a single neuron using the GCU activation function. Points in yellow were assigned a class label of +1 and points in blue were assigned a class label of -1 by a single GCU neuron.}
		   		     
		\end{figure}

		\begin{figure}[H]
		
		\center
		    \includegraphics[width = 0.5\textwidth]{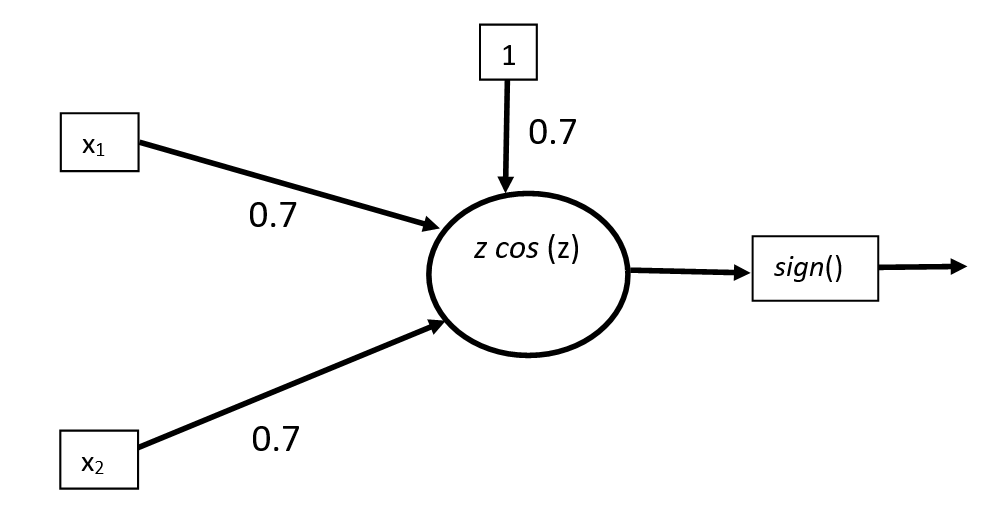}
		    \caption{A single neuron solution to the XOR problem. A single neuron with the GCU activation function is capable of learning the XOR dataset shown in (\ref{Eq: XOR dataset}) exactly. The signum function at the output is used to map output values to $\pm 1 $.  }
		    \label{Fig: XOR} 
		\end{figure}
		
		The XOR problem was solved by using a single neuron with oscillatory activation functions, mean-square loss and simple Stochastic Gradient Descent (SGD). A learning rate of $\alpha = 0.01$ and the SGD update rule $ \Delta \textbf{w} = \alpha (y - g(z))g'(z)\textbf{x}, \hspace{0.5cm} z = \textbf{w}^T\textbf{x} + b $ was used. The initial weight vector was initialized with uniform random  numbers in the interval $[-1 , 1]$.
		
		\vspace{0.5cm}
		
		The XOR function was successfully learned by a single neuron with the activation functions $g(z)$ chosen to be $g(z) = C(z) = z \cdot  \cos (z)$ and $g(z) = P_c(z) = z-\frac{z^3}{2}$ respectively.  The target $y$ for each input was taken as the class label namely 1 or -1. After training the output of the neuron is mapped to the class label in the usual manner. That is we assign positive outputs a label of +1 and negative outputs a label of -1.  This can be done simply by defining the class label $ Class(\textbf{x})$ for each input \textbf{x} to be  $Class(\textbf{x})= sign (g(\textbf{w}^T\textbf{x}+b)) $. Where the signum function is defined as 
		 $$  
		 sign\,(z) = 
		     \begin{cases}
		       1  &\quad \text{if}\quad z>0\\
		       -1 &\quad \text{if}\quad z<0\\
		       0  &\quad \text{if}\quad z= 0\\
		     \end{cases}
		$$

\textit{Definition 1}: The decision boundary of a single neuron is the set $ B = \{\textbf{x} \in \mathbb{R}^n:  g(\textbf{w}^T\textbf{x}+b) = 0 \}$. Where $g$ is the activation function.
		\vspace{0.5cm}
		
		That is the boundary is the set of inputs that elicit an output of zero from the neuron. Inputs corresponding to positive outputs are assigned the positive class bel (+1) and inputs corresponding to negative outputs are assigned the negative class label (-1) in accordance with $ Class(\textbf{x})= sign (g(\textbf{w}^T\textbf{x}+b)) $) as already discussed.
		
		\vspace{0.5cm}
		
		It is clear from \textit{Definition 1} that the decision boundary for any neuron that uses an activation function satisfying the condition $$ g(z) = 0 \Longleftrightarrow z = 0 $$ is $$ z = \textbf{w}^T\textbf{x}+b = 0 $$ In other words the decision boundary is a single hyperplane ($ B = H =  \{\textbf{x} \in R^n :  \textbf{w}^T\textbf{x}+b = 0 \}$).
		\vspace{0.5cm}
		
		However if $g$ is an oscillatory function like $C(z) = z\cos z $, the decision boundary is the set $ \{\textbf{x} \in R^n :  \text{GCU}(\textbf{w}^T\textbf{x}+b) = 0 \}$ and consists of infinitely many hyperplanes in the input space, since $C(z)$ has infinitely many roots. In particular the decision boundary with the GCU activation is a set of uniformly spaced parallel hyperplanes $ \textbf{w}^T\textbf{x}+b = \frac{\pi}{2} + n\pi $. Thus the input space is divided into parallel strips separated by the hyperplanes and point in adjacent strips are assigned different classes alternately. These parallel strips can be seen in the solution to the XOR problem (Fig. \ref{Fig: XOR}).
		
		\subsection{Characterization of Activation Functions} \label{limitations}
		 In the following, we prove that no single neuron with a strictly monotone activation function can learn the XOR function.
		
		We adopt the following notation: The output (activation) of a single neuron is given by $a = g(\textbf{w}^T\textbf{x}+b)$, where $g$ is the activation function. The hyperplane boundary associated with a neuron is the set of points: $$H = \{\textbf{x} \in \mathbb{R}^n: \textbf{w}^T\textbf{x}+b =0 \}$$
		
		The positive and negative half spaces are similarly defined to be:
		$$H_- = \{\textbf{x} \in \mathbb{R}^n: \textbf{w}^T\textbf{x}+b < 0 \}$$
		$$H_+ = \{\textbf{x} \in \mathbb{R}^n: \textbf{w}^T\textbf{x}+b > 0 \}$$

		Any hyperplane divides the input space $\mathbb{R}^n$ into 3 connected regions: the positive half-space $H_+$, the negative half-space $H_-$ and an affine-space $H$. The weight vector \textbf{w} points into the positive half-space $H_+$. 
		
		The distance between a point \textbf{x} and the hyperplane decision boundary $H$ is given by:  $$d( \textbf{x} ,H) =   \frac{| \textbf{w}^T\textbf{x}+b | }{ \| w \|} $$ .

		\textbf{Proposition 1:} Consider a single neuron with weight vector \textbf{w} and bias $b$ using an activation function that is monotonically strictly increasing with $ g(0) = 0 $. The class label assigned to an input $\textbf{x}$ by this neuron is defined to be $ C(\textbf{x})= sign (\textbf{w}^T\textbf{x}+b) $. If a point $\textbf{x}^1$ assigned to a particular class is at a distance $d_1$ from the hyperplane $ H = \{\textbf{x} \in \mathbb{R}^n: \textbf{w}^T\textbf{x}+b =0 \}$, then any other point $\textbf{x}^2$ at a distance $d_2 > d_1$ in the same halfspace as $\textbf{x}^1$ will be assigned to the same class by this neuron.
		
		\vspace{0.5cm}
		
		\textbf{Proof:}
		
		Case 1: Consider the case where $\textbf{x}^1 \in H_+$ AND $\textbf{x}^2 \in H_+$
		
		By assumption, $ z_1  = \textbf{w}^T\text{x}^1 + b > 0 $ AND $ z_2  = \textbf{w}^T\text{x}^2 + b > 0 $. Also $ d_1 < d_2 $
		
		Using the formula for $d(\textbf{x,H})$:
		
		$$ \frac{|\textbf{w}^T\textbf{x}^1+b|}{ \| w \| } < \frac{| \textbf{w}^T\textbf{x}^2+b|}{ \| w \| } $$
		
		$$ \frac{\textbf{w}^T\textbf{x}^1+b}{ \| w \| } < \frac{ \textbf{w}^T\textbf{x}^2+b}{ \| w \| } $$
		
		$$ \textbf{w}^T\textbf{x}^1+b < \textbf{w}^T\textbf{x}^2+b$$
		
		$$ z_1 < z_2 $$
		
		$ 0 < z_1 $ by assumption, thus $ 0 < z_1 < z_2 $.
		
		Since $g$ is strictly increasing and $ g(0) = 0 $ : $ 0 < g(z_1) < g(z_2) $
		
		$$ g(z_2) > 0 $$
		$$ Class(\textbf{x}^2)= sign (\textbf{w}^T\textbf{x}^2+b)=sign(g(z_2)) = 1 $$
		Thus $ x_2 \in H_+ $ and hence $x_2$ belongs to the same class as $x_1$.
		
		Case 2:  Consider the case where $\textbf{x}^1 \in H_-$ AND $\textbf{x}^2 \in H_-$
		
		By assumption, $ z_1  = \textbf{w}^T\text{x}^1 + b < 0 $ AND $ z_2  = \textbf{w}^T\text{x}^2 + b < 0 $. Also $ d_1 < d_2 $
		
		Using the formula for $d(\textbf{x,H})$:
		
		$$ \frac{|\textbf{w}^T\textbf{x}^1+b|}{ \| w \| } < \frac{| \textbf{w}^T\textbf{x}^2+b|}{ \| w \| } $$
		
		$$ \frac{-(\textbf{w}^T\textbf{x}^1+b)}{ \| w \| } < \frac{-( \textbf{w}^T\textbf{x}^2+b)}{ \| w \| } $$
		
		$$ \textbf{w}^T\textbf{x}^1+b > \textbf{w}^T\textbf{x}^2+b$$
		
		$$ z_1 > z_2 $$
		
		$ 0 > z_1 $ by assumption, thus $ 0 > z_1 > z_2 $.
		Since $g$ is strictly increasing and $ g(0) = 0 $ : $ 0 > g(z_1) > g(z_2) $
		
		$$ 0 > g(z_2) $$
		$$ Class(\textbf{x}^2)= sign (\textbf{w}^T\textbf{x}^2+b)=sign(g(z_2)) = -1 $$
		Thus $ x_2 \in H_- $ and hence $x_2$ belongs to the same class as $x_1$.

		Thus it is clear from Proposition 1, that if a point is assigned a particular class, other points further away from the boundary are automatically assigned the same class by strictly monotonic activation functions. However oscillatory activation functions are not subject to this limitation and hence can learn the XOR classification with a single neuron. In the following we introduce the notion of sign-equivalence of activation functions and use this property to characterize limitations of popular activation functions.
		
		\vspace{0.5cm}

		\textit{Definition 2}: A function $f:X\rightarrow\mathbb{R}$ is said to be sign equivalent to a function $g:X\rightarrow\mathbb{R}$ iff $sign(f(x)) = sign(g(x))$ for all $x \in X$.
		
		\vspace{0.5cm}
		
		It is clear that sign equivalence is actually an Equivalence relation on the set of all real-valued functions on a set. Further we note that the set of functions $G = \{f:\mathbb{R} \rightarrow \mathbb{R}:\quad f(0)=0\}$ is a vector space. Also the subset of functions of G that are sign equivalent to $I(z)=z$ form a convex cone in G. 
		\vspace{1cm}
		
		\textbf{Proposition 2:} Consider a single neuron that uses an activation function that is sign equivalent to the identity function $I(z)=z$, that is $sign(g(z))=sign(z)$. If $\textbf{x}^1, \textbf{x}^2 \in H_+$, then $ Class(\textbf{x}^1) = Class(\textbf{x}^2) = 1$ and if $\textbf{x}^1, \textbf{x}^2 \in H_-$, then $ Class(\textbf{x}^1) = Class(\textbf{x}^2) = -1$.
		\vspace{0.5cm}
		
		\textbf{Proof:}
		
		Case 1: Let $sign(g(z))=sign(z)$ AND $\textbf{x}^1, \textbf{x}^2 \in H_+$
		\vspace{0.5cm}
		
		$\Rightarrow  z_1 = \textbf{w}^T\text{x}^1 + b > 0 $ AND $ z_2  = \textbf{w}^T\textbf{x}^2 + b > 0 $
		\vspace{0.5cm}
		
		$\Rightarrow Class(\textbf{x}^1) = sign(g(z_1)) = sign(z_1) = 1$ AND $Class(\textbf{x}^2) = sign(g(z_2)) =sign(z_2) = 1$
		\vspace{0.5cm}
		
		Thus $\textbf{x}^1$ and $\textbf{x}^2$ belong to the same class.
		\vspace{0.5cm}
		
		Case 2: Let $sign(g(z))=sign(z)$ AND $\textbf{x}^1, \textbf{x}^2 \in H_-$
		\vspace{0.5cm}
		
		$\Rightarrow z_1 = \textbf{w}^T\text{x}^1 + b < 0 $ AND $ z_2  = \textbf{w}^T\textbf{x}^2 + b < 0 $
		\vspace{0.5cm}
		
		$\Rightarrow Class(\textbf{x}^1) = sign(g(z_1)) = sign(z_1) = -1$ AND $Class(\textbf{x}^2) = sign(g(z_2)) = sign(z_2) = -1$
		\vspace{0.5cm}
		
		Thus $\textbf{x}^1$ and $\textbf{x}^2$ belong to the same class.
		
		From Proposition 2 it is clear that a single neuron using the Swish activation function cannot solve the XOR problem. 
		\vspace{0.5cm}
		
		The Swish activation $ S(z) = \frac{z}{1+exp(-z)} $
		
		It is clear that $sign(S(z)) = sign(z)$ (since $\frac{1}{1+exp(-z)} > 0$).
		\vspace{0.5cm}
		
		Similarly a single neuron using the Mish activation function cannot solve the XOR problem. 
		\vspace{0.5cm}
		The Mish activation $ M(z) = z \tanh (\log (1+\exp (z))) $,
		\vspace{0.5cm}
		It is clear that $sign(M(z)) = sign(z)$ (since $\tanh (\log (1+\exp (z))) > 0$).
		\vspace{0.5cm}
		
		Based on propositions 1 and 2, single neurons that use monotonic activation functions and activation functions that are sign equivalent to $I(z)=z$ cannot learn the XOR function. To solve the XOR problem with a single neuron we must search for an activation that violates both the above conditions. In our work, the oscillatory function $ C(z) = z \cos z $ that violates both the above conditions is proposed and used to solve the XOR problem with a single neuron.
		In the following it is shown that although the GCU activation allows single neurons to the learn the XOR function, it is only slightly more computationally costly than Leaky ReLU. Also, the GCU activation is shown to be computationally cheaper than the recently popular Swish and Mish activation functions.

		\section{Comparison of Computational complexity for activation functions}
		
Table \ref{Table: Definitions} presents the definition of a variety of popular activation functions considered in this paper. From the definition it is clear that sigmoids, Swish, Mish and GCU are infinitely differentiable for all inputs, while ReLU and Leaky ReLU are differentiable everywhere except at zero.

\begin{table}[H]
		\centering
		\def\arraystretch{2}
		\begin{tabular}{c|c}
		\hline
		\textbf{Name}                & \textbf{Function}                     \\ \hline
		Logistic-sigmoid             & $\sigma(z) = \frac{1}{1+\exp{(-z)}}$  \\ %\hline
		tan-sigmoid & $T(z) = \tanh{z} = \frac{\exp (z) - \exp (-z)}{\exp (z) + \exp (-z)}$ \\ %\hline
		Rectified Linear Unit (ReLU) & $ R(z) = 
		      \begin{cases}
		       z  &\quad \text{if}\quad z>0\\
		       0 &\quad \text{if}\quad z<0\\
		      \end{cases} $                        \\ %\hline
		Leaky ReLU                   & $ L(z) = 
		      \begin{cases}
		       z  &\quad \text{if}\quad z>0\\
		       0.01 z &\quad \text{if}\quad z<0\\
		      \end{cases} $                                \\ %\hline
		Swish   & $ S(z) = \frac{z}{1+\exp{(-z)}}$        \\ %\hline
		Mish                         & $ M(z) = z \tanh{(\log (1+\exp{(z)}))}$ \\ %\hline
		Growing Cosine Unit (GCU)    & $ C(z) = z \cos{z}$                   \\ \hline
		\end{tabular}
		 \vspace{0.5cm}
		
		\caption{A list of activation functions considered in this paper and their definitions.}
		\label{Table: Definitions}
		\end{table}

\begin{figure}[H]
		 
\centering

\includegraphics[width= 10 cm]{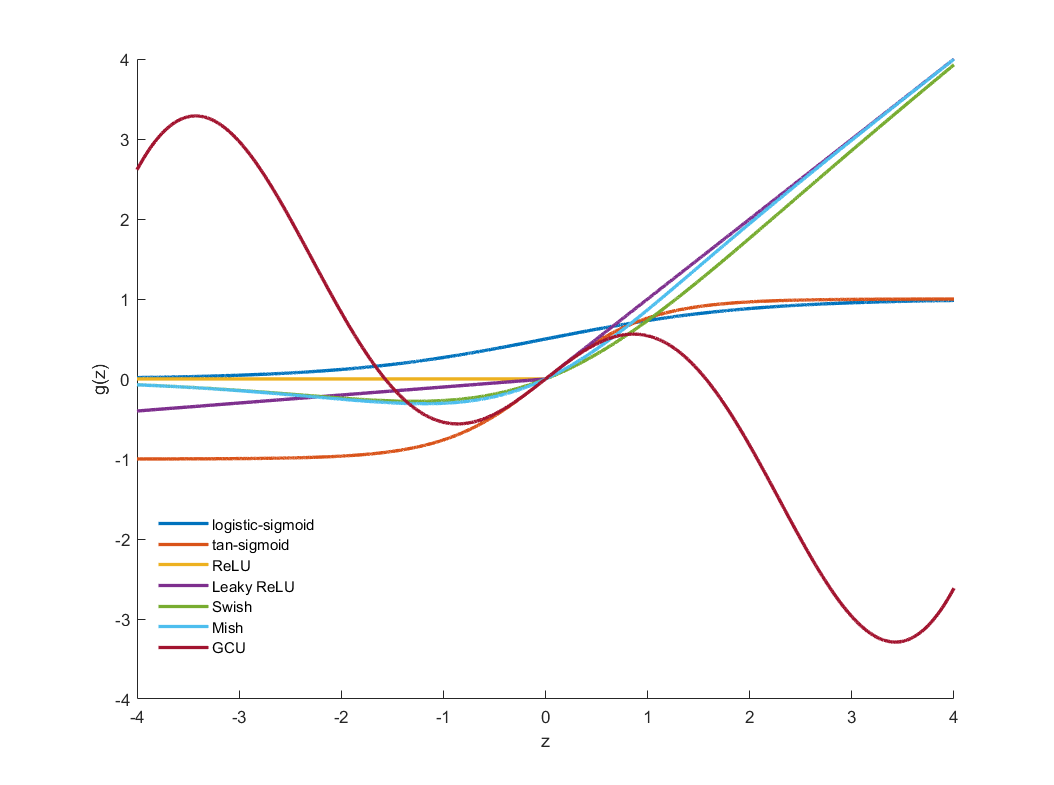}	
\caption{Plot of different activation functions}
\label{Fig: activations}

\includegraphics[width= 10 cm]{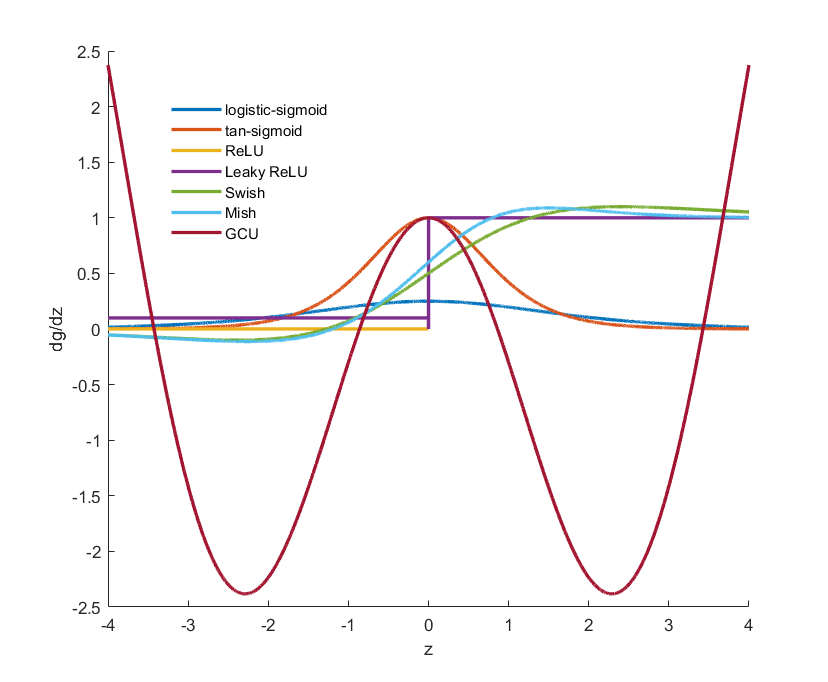}
\caption{Plot of first derivative of different activation functions}
\label{Fig: derivative activations}

\end{figure}

	Figures \ref{Fig: activations} and \ref{Fig: derivative activations} compare the features of different activation functions. It is clear that $C(z)$ and the other activation functions are very close to $I(z)=z$ for small values of z. This is desirable and has a regularizing effect since the network behaves like a linear classifier when initialized with small weights. During training the weights get updated and the nonlinear range of GCU is utilized as needed. In particular a GCU network can serve as a linear classifier if necessary avoiding overfitting effects. Also, the GCU activation temporarily saturates close to its first maximum and minimum values and mimics the behavior of sigmoids. For larger inputs GCU oscillates and is an unbounded function.
	 
\begin{figure}[H]
		\center
		    \includegraphics[width=8 cm]{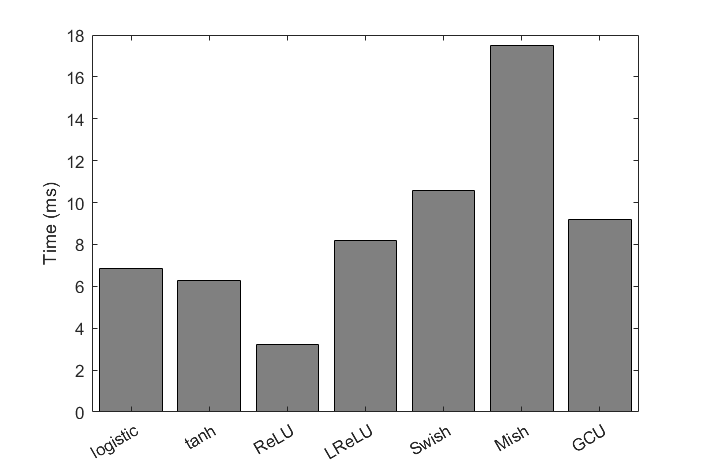}
		    \caption{Average time over 1000 independent runs. Where each run consisted of applying the activation to a vector of length $10^{6}$ with elements uniformly distributed in the interval [-5 , 5].}
		    \label{Fig: Time_Complexity}
		\end{figure}

	 It is evident from Table \ref{Table: Definitions} that the proposed GCU activation function is computationally cheaper than Swish and Mish activation functions. For example, GCU uses one transcendental function call and one multiplication whereas the Mish activation function uses 3 transcendental function calls and one multiplication. Computational experiments (Fig. \ref{Fig: Time_Complexity}) also clearly demonstrate that the proposed GCU activation is computationally cheaper than the popular state-of-the-art Swish and Mish activation functions.

\section{Comparison of performance on benchmark datasets}
\label{section:results}

		In the following a comparison of the performance of CNN models with different activation functions on the CIFAR-10 \cite{krizhevsky2009learning}, CIFAR-100, and Imagenette  \cite{imagenette} datasets is presented. CIFAR-10 consists of 60,000 $32 \times 32$ color images belonging to 10 classes with 6000 images per class with 5000 training and 1000 testing images per class. CIFAR-100 consists of 60,000 $32 \times 32$ color images belonging to 100 classes with 600 images per class with 500 training and 100 testing images per class. Imagenette is a subset of  ImageNet \cite{5206848} and consists of ten classes of easily recognized objects. The RMSprop optimizer \cite{Tieleman2012} is used with the categorical cross entropy loss function(softmax classification head). Experiments on CIFAR-10, CIFAR-100 were carried out with an initial learning, decay rate of $10^{-4}, 10^{-6}$ respectively. For Imagenette, this was $10^{-6}$, with no decay. The Xavier Uniform initializer was used to initialize the weights of the kernel layers. Tables  show performance with the proposed GCU activation function and popular activation functions in the convolution layers. The GCU activation is used only for the convolution layers and not for the dense layers since the GCU activation is computationally costlier than the ReLU activation.

		The average accuracy and loss over 5 independent runs (each of 25 epochs) is considered to average out the variations caused by random initialization of weights. The average and standard deviation of accuracy and loss on the testing set is presented. A compact CNN architecture (descibed in the Appendix) was used to learn CIFAR-10 and CIFAR-100.  This architecture consists of 4 convolution layers followed by dense layers. The same CNN architecture used for CIFAR-10 was used for CIFAR-100 by replacing the 10 neuron final softmax layer with a 100 neuron softmax layer. On the ImageNette problem, the VGG-16 backbone described in \cite{DBLP:journals/corr/SimonyanZ14a} was used.

\begin{figure}[H]
\centering
\includegraphics[width = 6 cm]{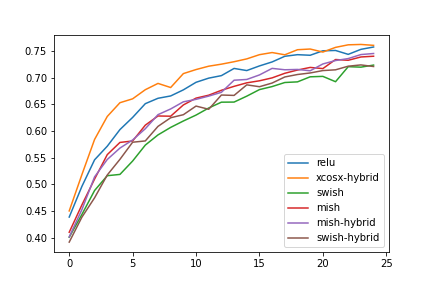} %
\includegraphics[width = 6 cm]{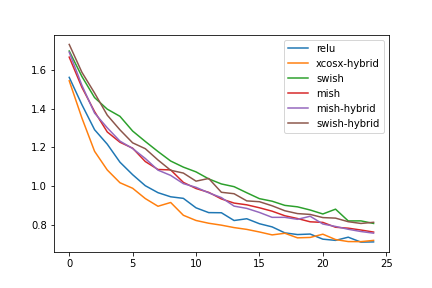}
\caption{Test dataset accuracy (left) and loss (right) on the CIFAR-10 dataset}
\label{cifar10_result}
\end{figure}

\begin{figure}[H]
\centering
\includegraphics[width = 6 cm]{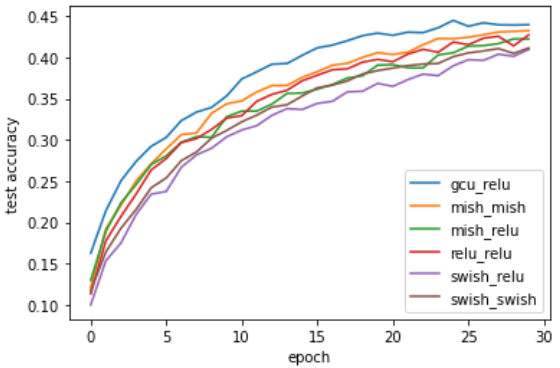} %
\includegraphics[width = 6 cm]{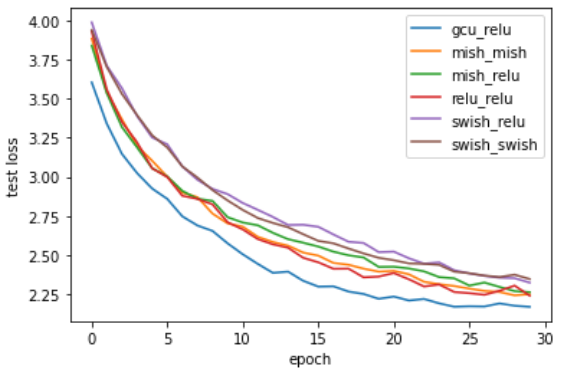}
\caption{Test dataset accuracy (left) and loss (right) on the CIFAR-100 dataset}
\label{cifar100_result}
\end{figure}

\begin{figure}[H]  
\centering
\includegraphics[width = 6 cm]{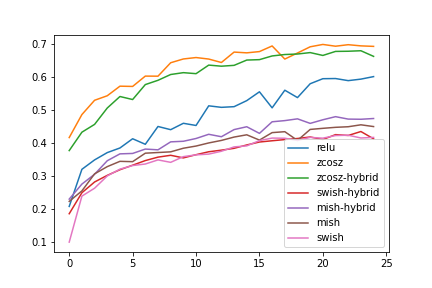}
\includegraphics[width = 6 cm]{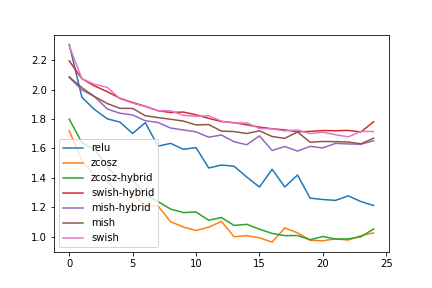}
\caption{Testing Accuracy (left) and Loss (right) on the Imagenette dataset}
		        
\label{Imagenette_result}
		    
\end{figure}

		% \usepackage{booktabs}
		\begin{table}[H]
		\centering
		\begin{tabular}{@{}l|l|r|r|r|r@{}}
		\toprule
		CONV. Layer & Dense Layer & Top - 1 Acc. \% & SD Acc. & Loss & SD Loss \\ \midrule
		ReLU  & ReLU  & 74.13 & 0.56 & 0.74 & 0.016 \\
		\textbf{GCU} & \textbf{ReLU}  & \textbf{75.64} & \textbf{0.47} & \textbf{0.73}              & \textbf{0.004} \\
		 Swish & Swish & 71.74 & 0.48 & 0.82              & 0.014 \\
		 Swish & ReLU  & 71.70  & 1.05 & 0.84                     & 0.016 \\
		 Mish  & Mish  & 74.22 & 0.62 & 0.77                     & 0.004 \\
		 Mish  & ReLU  & 73.20  & 0.74 & 0.79                     & 0.011 \\ \bottomrule
		\end{tabular}
		
		\caption{Test dataset accuracy with various activation functions on the CIFAR-10 dataset}
		\label{Table: cifar-10}
		\end{table}
		
\vspace{1 cm}

		\begin{table}[H]
		\centering
		\begin{tabular}{@{}l|l|r| r| r| r@{}}
		\toprule
		CONV. Layer &
		  Dense Layer &
		  Top- 1 Acc. \% &
		  SD Acc. &
		  Loss &
		  SD Loss \\ \midrule
		 ReLU  & ReLU  & 41.29                         & 0.43  & 2.31 & 0.016 \\
		\textbf{GCU} & \textbf{ReLU}   & \textbf{43.42}                   & \textbf{0.36}  & \textbf{2.23} & \textbf{0.004} \\
		 Swish & Swish & 39.37                  & 0.40  & 2.43 & 0.014 \\
		 Swish & ReLU  & 38.46                         & 0.42  & 2.45 & 0.016 \\
		 Mish  & Mish  & 41.13                         & 0.36 & 2.33 & 0.004 \\
		 Mish  & ReLU  & 39.83                         & 0.37  & 2.39 & 0.011 \\ \bottomrule
		\end{tabular}
		\vspace{0.5cm}
		\caption{Test dataset accuracy with various activation functions on the CIFAR-100 dataset}
		\label{Table: cifar-100}
		\end{table}
		
		% \usepackage{booktabs}
		\begin{table}
		\begin{tabular}{@{}l|l|l|l|r|r@{}}
		\toprule
		Convolution Layer & Activation Dense Layer & Top- 1 Acc. \% & SD Acc.       & Loss & SD Loss \\ 
		\hline
		ReLU  & ReLU  & 60.28 & 0.60 & 1.21 & 0.02 \\ 
		\textbf{GCU}      & \textbf{ReLU}          & \textbf{68.27} & \textbf{1.01} & \textbf{1.00}             & \textbf{0.03}                \\ 
		GCU   & GCU   & 67.87 & 0.37 & 1.07 & 0.02 \\ 
		Swish & Swish & 43.02 & 0.65 & 1.69 & 0.01 \\ 
		Swish & ReLU  & 42.96 & 0.27 & 1.71 & 0.03 \\ 
		Mish  & Mish  & 48.72 & 1.79 & 1.56 & 0.06 \\ 
		Mish  & ReLU  & 44.32 & 2.16 & 1.84 & 0.13 \\ \bottomrule
		\end{tabular}
		\vspace{0.5cm}
		\caption{Test dataset accuracy with various activation functions on the Imagenette dataset}
		\label{Table: imagenette}
		\end{table}

		Tables \ref{Table: cifar-10}, \ref{Table: cifar-100} and \ref{Table: imagenette} show that the use of the GCU activation in the convolution layers provides the best performance among all architectures considered. This is particularly evident on the VGG-16 network trained on the Imagenette dataset, where the GCU models outperform all ReLU architectures by 7\%. The models with GCU in the convolution layers also converge faster during training as highlighted by Figs \ref{cifar10_result}, \ref{cifar100_result} and \ref{Imagenette_result}.
		
		\subsection{Visualization of learnt filters}
		
\begin{figure}[H]
		    \includegraphics[width=6 cm]{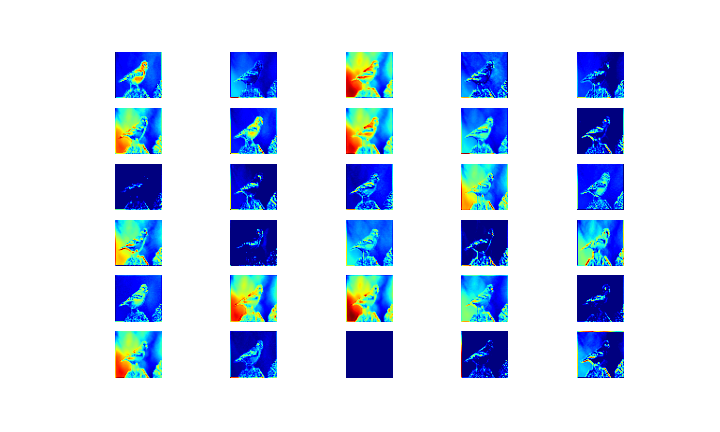}
		    \includegraphics[width=6 cm]{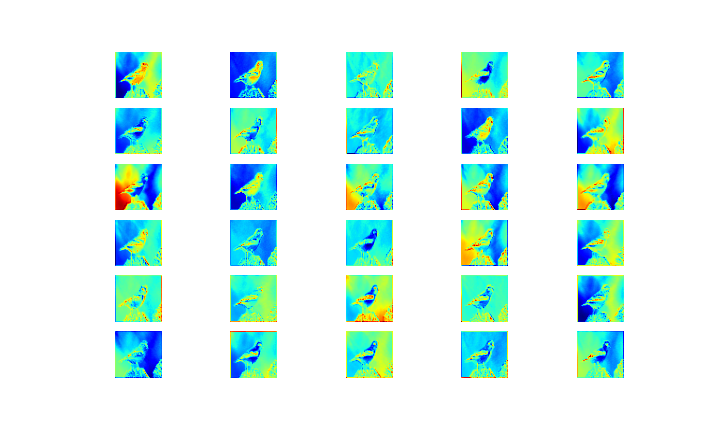}
		    \caption{Comparisons of filter output in Layer 3 with ReLU and GCU activation functions. Outputs from ReLU filters are shown in the 5 leftmost columns while the 5 rightmost columns show the outputs of filters with the GCU activation.}
		        \label{cifar10_filters_1}
		\end{figure}

\begin{figure}[H]
		    \includegraphics[width=6 cm]{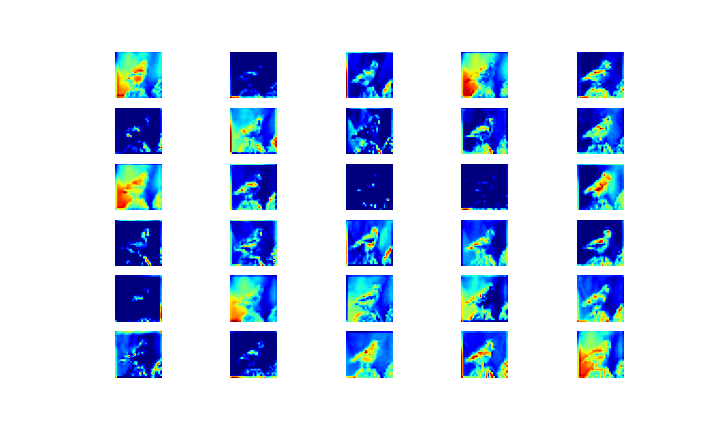}
		    \includegraphics[width=6 cm]{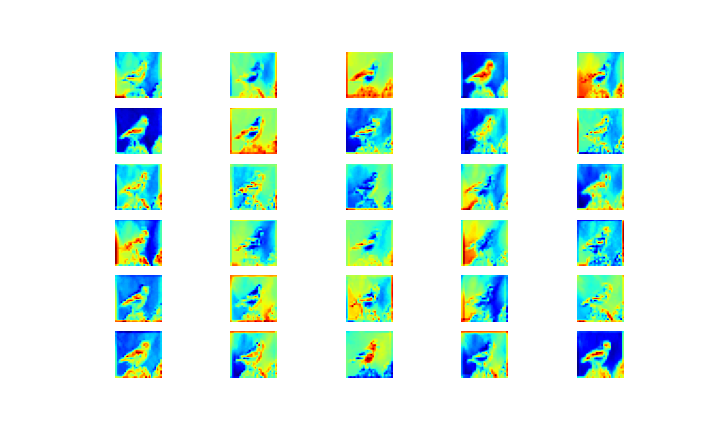}
		    \caption{Comparisons of filter output in Layer 6 with ReLU and GCU activation functions. Outputs from ReLU filters are shown in the 5 leftmost columns while the 5 rightmost columns show the outputs of filters with the GCU activation.}
		        \label{cifar10_filters_2}
		\end{figure}

\begin{figure}[H]
		    \includegraphics[width=6 cm]{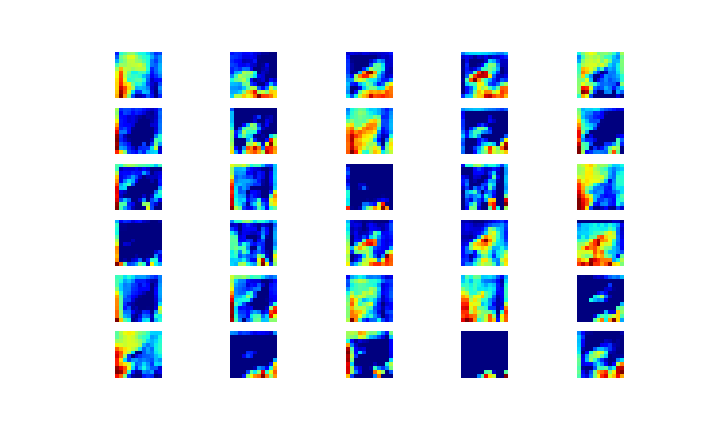}
		    \includegraphics[width=6 cm]{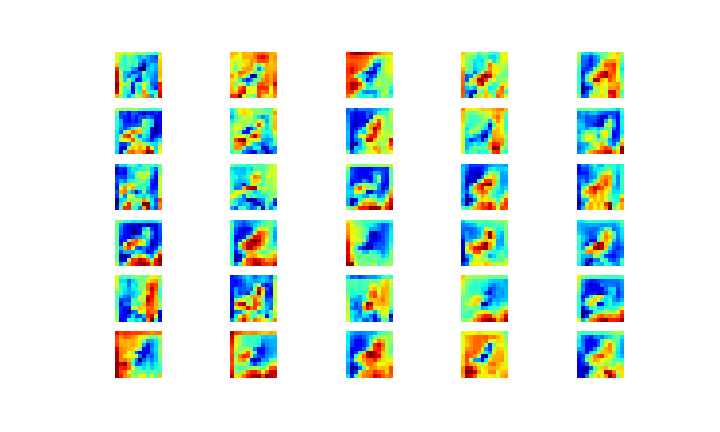}
		    \caption{Comparisons of filter output in Layer 10 with ReLU and GCU activation functions. Outputs from ReLU filters are shown in the 5 leftmost columns while the 5 rightmost columns show the outputs of filters with the GCU activation.}
		        \label{cifar10_filters_3}
		\end{figure}

		  Figs. \ref{cifar10_filters_1}, \ref{cifar10_filters_2} and \ref{cifar10_filters_3} present the activation of the filters in successive layers  for the ImageNette dataset. It is clear from Figs. \ref{cifar10_filters_1}, \ref{cifar10_filters_2} and \ref{cifar10_filters_3} that both ReLU and GCU convolution layers hierarchically detect the features of a bird in the input image. However, it is quite clear that the feature detectors with GCU activation function are more confident and correspond to larger outputs (red pixels correspond to larger values). In particular the 5 rightmost columns in Fig. \ref{cifar10_filters_3} clearly show the detection of the bird image in red. Thus, it appears that convolutional filters with GCU activation are able to segment and detect the bird image significantly more clearly and accurately than with the ReLU activation function. These filter output visualizations qualitatively confirm the quantitative higher accuracy results with the GCU activation shown in Table \ref{Table: imagenette}.
		
		\subsection{Effect of activation function on gradient flow}
		The ReLU has derivative equal to zero for all negative values. In contrast the Leaky ReLU activation function has a fixed non-zero derivative value for  negative inputs resulting in faster learning. However the Leaky ReLU activation also saturates for large negative inputs. In contrast to ReLU and sigmoidal activation functions, the GCU activation function never saturates.
		For small values the GCU activation behaves like the linear activation function and has a derivative value close to 1, since $z\cos(z) \approx z$. This linear behavior (with derivative close to 1) of the GCU activation close to the origin leads top faster training in the beginning when the neural network parameters are initialized with small random values. For larger values the derivative of the GCU activation decreases and the GCU activation mimics the tan-sigmoidal activation function in the interval $[-\frac{\pi}{2},\frac{\pi}{2}]$. Beyond this interval the GCU oscillates and every zero of the GCU activation corresponds to a hyperplane decision boundary.
		
\begin{figure}[H]
		\centering
		
		    \includegraphics[width = 8 cm]{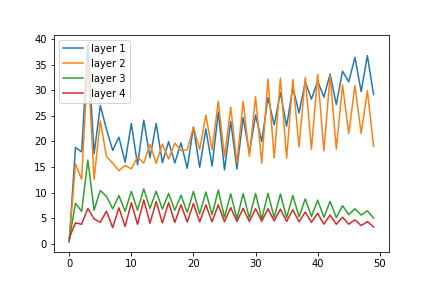}
		    \caption{Variation of RMS layer gradients with training epochs with the ReLU activation functions.}
		    \label{relu}
		    \includegraphics[width = 8 cm]{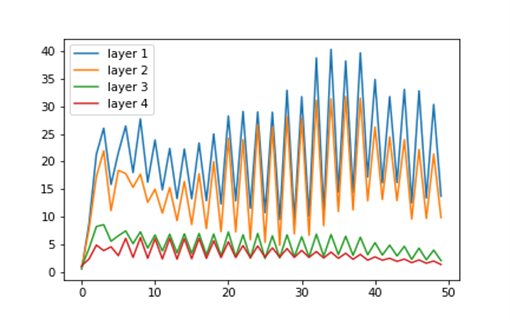}
		    \caption{Variation of RMS layer gradients with training epochs with the Leaky ReLU activation functions.}
		    \label{Lrelu}
		    \end{figure}

\begin{figure}[H]
		    \centering
		    \includegraphics[width = 8 cm]{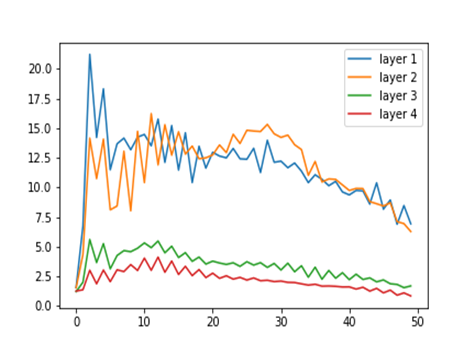}
		    \caption{Variation of RMS layer gradients with training epochs with the GCU activation functions.}
		    \label{GCU}
		    \end{figure}

	In figures \ref{relu}, \ref{Lrelu} and \ref{GCU} the Root Mean Square(RMS) value of the gradients for every convolution layer in the model for each activation function is plotted against every 400 mini-batches. The RMS value of the gradients for ReLU is in the range of 0 and 40. In general, the change in gradients is high and the gradient values oscillate wildly. In the 4th layer  the gradients approach 0 but do not reach it. Similar to ReLU, the RMS value of the gradients of each layer for Leaky ReLU is in the range 0 to 40.  The gradient values for Leaky ReLU also oscillate wildly and are similar to the gradient values with ReLU.		
		
		 The RMS values of gradients of each layer with the GCU activation function is in the range 0 to 20 . It is observed that the oscillations in the gradient values with GCU activation result is significantly reduced compared to ReLU and Leaky ReLU. The reduced oscillations of the gradient values with GCU activation might explain the faster training of networks observed with the GCU activation.
		
		 \section{Conclusion}
		This paper explored the possible advantages of using oscillatory activation functions that differ drastically from ReLU like activation functions in the convolution layers of deep CNNs. In the past some weakly non-monotonic activations functions that very closely resemble ReLU have been considered \cite{DBLP:journals/corr/abs-1908-08681}, but these activation functions suffer from the same theoretical limitations as ReLU like activation functions (Section \ref{limitations}). Extensive comparisons of performance on CIFAR-10, CIFAR-100 and Imagenette indicate that a new activation function $C(z)= z \cos (z) $ significantly outperforms all popular activation functions on testing-set accuracy and speed of convergence. The proposed GCU activation function allows certain classification tasks to be solved with significantly fewer neurons. In particular the famous XOR problem which hitherto required a network with a minimum of 3 neurons for its solution was solved with a single GCU neuron. Intriguingly the decision boundary of a single GCU neuron is observed to consist of infinitely many parallel hyperplanes instead of a single hyperplane since the GCU activation has infinitely many zeros. Experimental results indicate that the use of oscillatory activation functions improve gradient flow and alleviate the vanishing gradient problem. Improved gradient flow can be attributed to GCU activation having small derivative values only close to isolated points in the domain instead of on entire infinite intervals. The findings in this research indicate that a wider class of functions that drastically differ from the popular ReLU like functions can serve as useful activation functions in CNNs. In particular the recent discovery of neurons with oscillating activation in the human cerebral cortex capable of individually learning the XOR function like the GCU neuron proposed in this paper provides a biological inspiration for oscillating activation functions \cite{doi:10.1126/science.aax6239}, \cite{poirazi2003pyramidal}. Future work will explore more oscillating activation functions to attempt to identify even better activation functions \cite{DBLP:journals/corr/abs-2111-04020}. 
		
\vspace{15 cm}		
		
\appendix{Appendix: Compact CNN architecture used to solve CIFAR-10}
\begin{figure}[H]
\centering

\includegraphics[height = 18 cm]{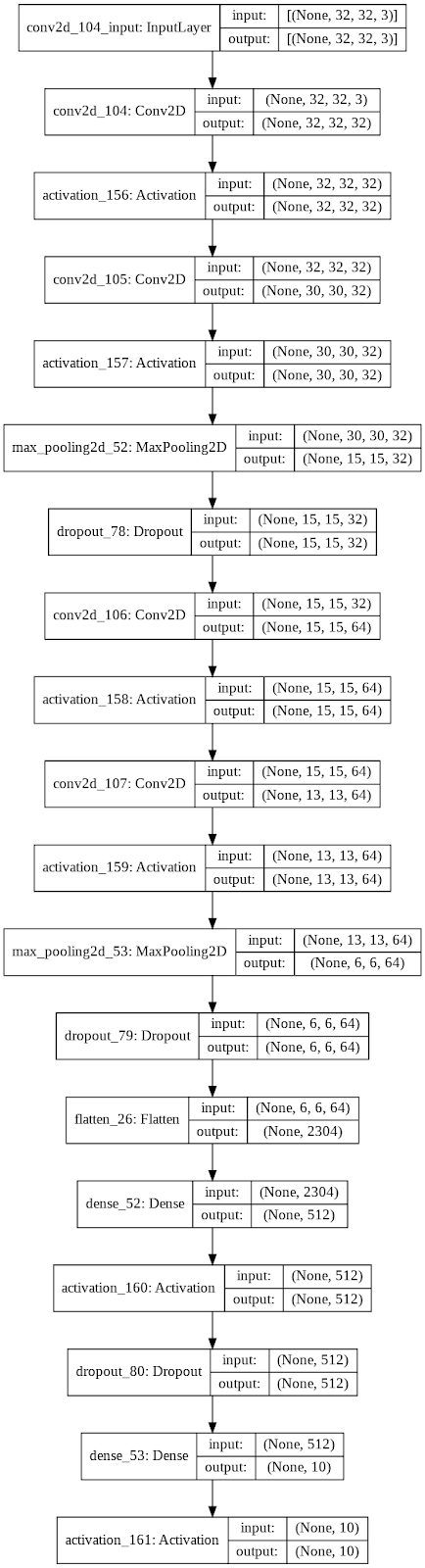}
		
\end{figure}

		
		
		
		\bibliographystyle{elsarticle-num}
		\bibliography{bibliography}

		
		\end{document}